\documentclass[11pt]{article}
\usepackage{coling2020}
\usepackage{times}
\usepackage{url}
\usepackage{latexsym}
\usepackage{hyperref}
\usepackage[toc,page]{appendix}
\usepackage{subcaption}
\usepackage{linguex}
\usepackage{coptic}
\usepackage{boldline}
\usepackage{enumitem}
\usepackage[skip=1pt]{caption}
\usepackage{graphicx} 
\usepackage[export]{adjustbox} 

\colingfinalcopy 

\title{Exhaustive Entity Recognition for Coptic: Challenges and Solutions}

\author{Amir Zeldes \\
  Georgetown University \\
  {\tt amir.zeldes@georgetown.edu} \\\And
  Lance Martin \\
  Catholic University of America \\
  {\tt 71martin@cua.edu} \\\AND
  Sichang Tu \\
  Georgetown University \\
  {\tt st1018@georgetown.edu} \\}

\date{}

\begin{document}
\maketitle
\begin{abstract}
 Entity recognition provides semantic access to ancient materials in the Digital Humanities: it exposes  people and places of interest in texts that cannot be read exhaustively, facilitates linking resources and can provide a window into text contents, even for texts with no translations. In this paper we present entity recognition for Coptic, the language of Hellenistic era Egypt. We evaluate NLP approaches to the task and lay out difficulties in applying them to a low-resource, morphologically complex language. We present solutions for named and non-named nested entity recognition and semi-automatic entity linking to Wikipedia, relying on robust dependency parsing, feature-based CRF models, and hand-crafted knowledge base resources, enabling high accuracy NER with orders of magnitude less data than those used for high resource languages. The results suggest avenues for research on other languages in similar settings.

\end{abstract}

\section{Introduction}

Recent developments in high quality NLP have been likened to a tsunami \cite{Manning2015}, powered largely by Big Data for tasks such as Named Entity Recognition (NER), and continuous meaning representations in the form of word embeddings, for English and other languages \cite{Upadhyay2016,PetersNeumannIyyerEtAl2018}. Meanwhile, low resource and historical languages have not been able to take advantage of these advances for several reasons: 1.~Gold datasets for most tasks are much smaller -- English OntoNotes 5.0 NER \cite{HovyMarcusPalmerEtAl2006} has 1.7M words, more than all digitized text available in many ancient languages; 2.~Work on embeddings often assumes at least $\sim$30 million words of training data \cite{CaoEtAl2018};\footnote{This limitation has also been recognized in NER work targeting Latin \cite{ErdmannEtAl2016}, where models based on word embeddings did not yield competitive results.} 3.~For historical and non-standardized languages, orthographic variation, regional differences, lacunae and other phenomena make learning from unlabeled data exceptionally hard.

In this paper, we report on a series of experiments, including successes and failures, in creating historical language resources for the Coptic language. The main contributions of this work are threefold: 
\begin{enumerate}[itemsep=0pt]
    \item We provide a new dataset annotated for entity types, named and non-named, nested entities, and entity linking, i.e. connecting spans of text to a table of specific people, places and other identifiers.
    \item We evaluate recent NLP approaches to NER in a low resource setting and show that they do not perform adequately for the needs of (Digital) Humanists.
    \item We present an alternative approach relying on dependency parsing, feature-based CRF models and a modest sized knowledge base, which are less labor intensive to produce than million-word training datasets, and offer a way forward for high accuracy NER in a morphologically complex, low resource historical language.
\end{enumerate}

\paragraph{The Coptic language} Coptic is the last stage of Ancient Egyptian, written in the first millenium CE in a modified Greek alphabet with additional letters derived from Egyptian Demotic script. Coptic writings are abundant, ranging from religious writings, such as hagiographies, homilies, magical texts, and ascetical treatises to social documents including letters, legal documents, and administrative records \cite{Bagnall2009,Fournet2020}. A vast amount of original compositions survive next to translated texts, especially of religious writing. Many early monastic centers wrote extensively in Coptic and many Manichaean and gnostic works survive only or primarily in Coptic. Although it was rarely the most common bureaucratic language, Coptic papyri provide crucial information about official and economic affairs in Egypt during Roman, Byzantine, and Arab rule. As such, Coptic literature is important to general history and the history of Christianity and other religions in and outside Egypt in antiquity. 

Coptic grammar presents challenges to automated processing, including agglutinative morphology (fusion of multiple affixes to content words, as in \ref{ex:agglut}), incorporation (e.g.~compound verbs which contain fused verbal arguments, as in \ref{ex:incorp}), and spelling variation which characterizes many ancient texts, including for widespread Greek loanwords \ref{ex:normalization}.

\ex. 	\begin{coptic}ne\0a\0c\0tre\0f\0cwtm\end{coptic} \quad\quad \textit{ne-a-s-tre-f-sōtm} \quad `she had made him hear' \\
	\small{PRET-PST-she-CAUS-he-hear}\label{ex:agglut} 
	
\ex. \begin{coptic}h1etb\0psuxh\end{coptic} \quad  \textit{hetb-psych\={e}}  `(to) soul-kill' =  \textit{h\={o}tb} `kill' +  \textit{psych\={e}} `soul' \label{ex:incorp}

\ex.	\begin{coptic}kolluqe\end{coptic} \quad\quad~  \textit{kollyk$^y$e} `group', misspelled for  \textit{kollēgion} (Gr. form of Lat. \textit{collegium})\label{ex:normalization}

In \ref{ex:agglut}, the sequence corresponding to `she had made him hear' is spelled in Coptic without spaces (or hyphens), meaning segmentation is needed before individual words, and then entities, can be recognized. In \ref{ex:incorp}, a compound verb fuses to `soul', changing the form of the verb and again requiring splitting. In \ref{ex:normalization}, spelling variation makes the Greek \textit{kollēgion} hard to recognize.

These challenges are not the focus of this paper, and our results below will be based on normalized, gold segmented text. However they mean that in practice, entity recognition will degrade when applied to automatically segmented and normalized text. Since humanists often have very high standards of accuracy, we therefore require robust solutions, as well as possibilities of incorporating semi-automatic correction steps, which we discuss below.

\paragraph{Entity recognition} Entity annotation for DH studies can encompass three related but distinct tasks: 
\begin{itemize}[itemsep=0pt]
    \item  Narrow NER, which identifies words referring to \textit{named} entities, and classifies them as \textsc{per}son, \textsc{place}, \textsc{org}anization etc. Named entities are often assumed not to overlap, possibly creating awkward spans (e.g.~[UK]$_\textsc{place}$ [Prime Minister Boris Johnson]$_\textsc{per}$, in which the place span is outside of the person span, despite belonging to the same syntactic phrase).
    \item The more exhaustive task of Non-named/Nested Entity Recognition (NNER), including all spans referring to an entity, including overlapping entities (e.g.~[[[\textit{LA}]$_\textsc{place}$ \textit{police}]$_\textsc{org}$ \textit{chief}]$_\textsc{per}$).
    \item Entity Linking, sometimes referred to as Wikification, in which notable entities are associated with a unique identifier, often a corresponding article about the entity in Wikipedia. 
\end{itemize} 

For Coptic, we are interested in all three, since texts include unnamed people and places of interest (unspecified `monks', unnamed locations such as `the monastery') which scholars may want to find, count and compare in texts. For named entities (esp. people/places), identities are important but difficult to find with string searches, since names repeat frequently (`Johannes' can be John the Baptist, St. John the Evangelist, Apa Johannes the Archimandrite, etc.), and marking up unique identities allows linking resources to data from other projects and languages, increasing their utility and discoverability. In the following, we present a new dataset for Coptic (N)NER and Wikification (Section \ref{sec:data}), test out-of-the-box and custom approaches to our tasks (Section \ref{sec:experiments}) and discuss some applications for humanities research using entities (Section \ref{sec:applications}). Section \ref{sec:conclusion} draws the conclusion and suggests directions for further study.



\section{Data}\label{sec:data}

\paragraph{Entity annotation} As an underlying dataset we chose the Coptic Treebank \cite{ZeldesAbrams2018} from the Universal Dependencies project (V2.6, \url{https://universaldependencies.org/}), which contains close to 50,000 genre-balanced tokens (30K train, and 10K each dev/test) annotated for gold dependency syntax, POS tags and lemmatization, covering both native and translated texts. For automatic annotation using the tools in the next section we use the larger dataset made available by Coptic Scriptorium (approx. 1M words, \url{http://copticscriptorium.org}) which we aim to make searchable for entity information. We annotate 10 entity types, shown in Table \ref{tab:entclasses} with their proportions in the corpus. Although some of the types, such as events, plants, organizations and animals, are comparatively rare, they are nevertheless distinct and potentially very interesting; for example, events cover turning points in stories, such as a person's death, a war or conquest, famine, etc., while organizations often identify factions in theological and military conflicts (the Catholic Church, Diocletian's army, etc.). When NER is applied at scale, we expect them to be useful in conducting research on the underlying entity types.

\begin{table*}[h!bt]
\begin{center}
\begin{tabular}{l r l|l r l}

entity type & \% & examples & entity type & \% & examples \\
\hline
\textsc{abstract} & 28.72 & `humility' & \textsc{person} & 39.92 & `all angels' \\
\textsc{animal} & 1.09 & `200 horses' & \textsc{place} & 10.87 & `Alexandria' \\
\textsc{event} & 2.00 & `his death' & \textsc{plant} & 1.00 & `wheat' \\
\textsc{object} & 9.79 & `bottles' & \textsc{substance} & 1.43 & `water' \\
\textsc{organization} & 0.86 & `the army' & \textsc{time} & 4.31 & `ten years' \\

\end{tabular}
\end{center}
\caption{\label{tab:entclasses} Entity types in our data with examples and percentages.}
        \par\vspace{-10pt}\par
\end{table*}

\begin{figure}[h!bt]
    \centering
    \includegraphics[width=0.5\textwidth,frame]{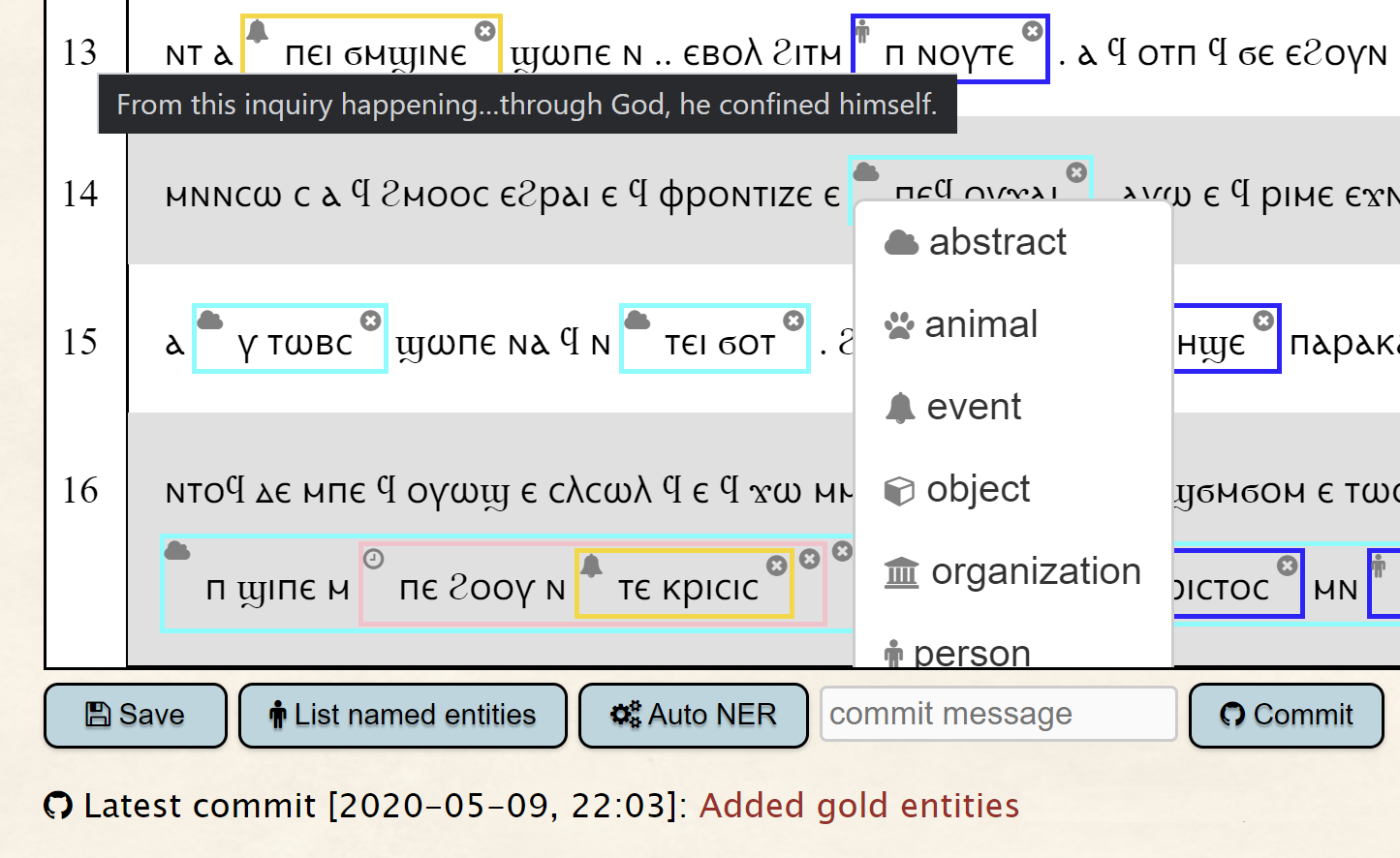}
    \caption{Annotation interface.}
    \label{fig:interface}
        \par\vspace{-10pt}\par
\end{figure}

As an annotation interface we use the version controlled online editor GitDox \cite{ZhangZeldes2017} , shown in Figure \ref{fig:interface}, which visualizes entity spans as boxes, color-coded for entity type, and enforces strict entity nesting (entities only overlap fully contained entities), as well as no crossing of sentence boundaries (based on the Coptic Treebank's sentence splits). 

To assess the reliability of our annotations, we carried out an inter-annotator agreement study by double annotating 1,162 tokens, containing 147 entities after adjudication. The annotators both had college level training in Sahidic Coptic and previous experience annotating Coptic corpora for other categories, such as part-of-speech tagging. Since measuring agreement on nested entities is non-trivial due to overlapping entities, we compute several metrics: 

\begin{enumerate}[itemsep=0pt]
    \item Cohen's Kappa, based on a single gold label per token, such that each token reflects the category of its deepest nested span (e.g.~``for the army of Diocletian'' becomes [O, B-org, I-org, I-org, B-per], since an organization begins at word 2, and continues until a person begins at the last word). This results in 21 possible tags (2 * entity types + `O').
    \item Micro-averaged mutual precision, recall and f-score, taking each annotator as a control for the other. Precision and recall take the perspective of annotator \#1, but can equally be reversed.
    \item Head accuracy: ignoring spans, what proportion of head nouns are assigned correct types.
\end{enumerate}

The results are shown in  Table \ref{tab:agreement}, including scores for span agreement without entity types. Agreement is far above chance, with a typed and untyped kappa of .85 and .9 respectively, falling in the so-called `perfect' bracket of 0.8--1.0 \cite{LandisKoch1977}. Precision, recall and f-scores are harder to interpret, since neither annotator corresponds to a `ground truth'; however, since we use the same metrics to evaluate systems below, these values help give a ceiling for automatic accuracy in Section \ref{sec:experiments}. 

\begin{table*}[h!bt]
\begin{center}
\begin{tabular}{l r r r r r}
 & kappa & F1 & precision & recall & head acc \\
\hline
\textit{typed} & 0.859 & 0.807 & 0.785 & 0.829 & 0.903 \\
\textit{untyped} & 0.902 & 0.883 & 0.859 & 0.908 & N/A \\

\end{tabular}
\end{center}
\caption{\label{tab:agreement} Results of the inter-annotator agreement experiment.}
        \par\vspace{-10pt}\par
\end{table*}

Comparing typed and untyped metrics, we also see annotators agree much better on spans than on entity types. To understand why, we examine cases like \ref{ex:soil}--\ref{ex:ground}.

\exg. mn 	          [hah   n-kah] haro-u \quad\quad  `it has no [soil]$_\textsc{obj/subst}$ under them' \label{ex:soil} \\  
     \small{not} \small{much}  \small{of-earth}  \small{below-them}   \\

\exg. [nē]    ne               nt-a-u-jo-ou            ejm-p-kah        et-nanou-f   `[those]$_\textsc{per/plant}$ sown on the good ground'  \label{ex:ground} \\
     \small{those} \small{are}    \small{\textsc{rel}-\textsc{pst}-they-sow-them}  \small{upon-the-earth} \small{\textsc{rel}-good-it}  \\

In \ref{ex:soil}, annotators disagreed whether `earth' is a \textsc{substance} or concrete \textsc{object}, which is murky in context. In \ref{ex:ground}, as part of a parable in which Jesus likens people to plants, the \textsc{plant} in one analysis, is resolved as a \textsc{person} in the other. It seems likely that NLP tools will err in assigning the more common category to ambiguous words across such contexts. At the same time, the head-based metrics shows that annotators mostly agree on entity types when exact spans are ignored (about 90\% of the time). Most disagreements are due to entities omitted by one of the annotators, as in \ref{ex:birds}, where one annotator ignored `sky' in `birds of the sky' as non-referential, while another treated it as a \textsc{place}.

\exg. n-halate   n-[t-pe] \quad\quad `The birds of [the sky]$_\textsc{place}$' \\
     the-birds of-the-sky \label{ex:birds}\\
      
While not perfect, the analysis suggests that human-quality scores of around 90\% would provide a good basis for humanists' work using the entity annotations, with at least some of the disagreements revolving around weakly referential cases, which may be of less interest to researchers (see Section \ref{sec:applications}).

\paragraph{Entity Linking} For entity linking, we followed the broadly used approach of linking mentions of named entities to their corresponding Wikipedia articles \cite{MilneWitten2008,ShnaydermanEtAl2019}, i.e.~Wikification. Using Wikipedia as a table or authorities brings a number of advantages and disadvantages, though we feel the former outweigh the latter substantially. The main advantages are obtaining an existing high quality table of authorities, and the wide range of other projects using Wikipedia identifiers, including resources linked in multiple languages \cite{McNameeEtAl2011}. Many notable Coptic person entities are not indexed in other relevant inventories, but do have Wikipedia pages. At the same time, Wikipedia identifiers are also available in the largest subset of projects, including broad projects on antiquity, such as Pleiades, and targeted ones adjacent to ours, such as Syriaca.\footnote{For example \url{http://syriaca.org/place/572.html} and  \url{https://pleiades.stoa.org/places/727070} are both aligned with the Wikipedia entry for Alexandria, meaning their entries can easily be aligned with ours.}

Due to the high coverage of Wikipedia (especially for people) and the desirability of re-using common, existing identifiers, we opted to annotate our gold entity dataset with Wikipedia identifiers, which included 610 named entity types, of which 441 were found to have Wikipedia articles, amounting to 104 unique identities (i.e.~distinct articles). The remainder consisted of minor entities or unknown/unidentified people mentioned in texts, such as `Bibrus', an unidentified minor character in the Dormition of John, or Mahlon in the Book of Ruth. We evaluate the feasibility of using these seed annotations for automatic Wikification in Section \ref{sec:experiments}, and give plans for wikifying more data in Section \ref{sec:conclusion}.

\paragraph{Availability} Our annotations are made freely available online under a Creative Commons Attribution (CC-BY) 4.0 license, matching the license of the Coptic Treebank. To facilitate re-use and interoperability, data is versioned on GitHub in UD's CoNLL-U format, Corpus Workbench format \cite{Christ1994}, PAULA stand-off XML \cite{Dipper2005} and TEI XML (\url{https://tei-c.org/}). 

\section{Experiments in Automatic Entity Recognition}\label{sec:experiments}

\paragraph{Mention detection} Before we can evaluate entity classification and linking, we must first find entity span candidates in texts. As baselines we consider: a. using all and only nouns as entity spans (\textsc{noun}); and b. using all and only sequences of words attested as an entity in the training data (\textsc{lookup}). The \textsc{noun} strategy will only recall single token mentions, and include incorrect non-referring expressions. \textsc{lookup} should have few false positives, but low recall, since novel strings in the test data will be missed.

As competitive solutions we consider two families of methods: 1. In \textsc{parse}, entities are assumed to cover the spans of phrases headed by nouns, as identified by the syntax tree; 2. \textsc{sequence}: sequences of words in the corpus are scanned and classified as entities using a neural sequence tagger.  Since we have reference syntax trees for our data, we can evaluate performance with gold and predicted trees from an automatic parser. 

As a `sequence' based system, we train a state of the art NNER system \cite{YuEtAl2020}, which relies on a bidirectional recurrent neural network (RNN) with biaffine attention. The system scores spans based on start and end token indices, considering all possible spans, including nested mentions, and outputs a probability for each span category, or `no-entity' (=`O(outside)' in BIO encoding) for spans not predicted as mentions. The system relies on word embeddings, which we provide using Word2Vec \cite{MikolovEtAl2013} based on $\sim$1 million tokens of unannotated, automatically segmented Coptic text from Coptic Scriptorium \cite{ZeldesSchroeder2015} with a vocabulary size of $\sim$11,000 types and 50 dimensional representations.

\begin{table*}[h!bt]
\centering
\begin{tabular}{l|lll|lll}
 & \multicolumn{3}{c|}{\textbf{exact span match}} & \multicolumn{3}{c}{\textbf{fuzzy head span}} \\
\textbf{method} & \textbf{R} & \multicolumn{1}{c}{\textbf{P}} & \multicolumn{1}{c|}{\textbf{F1}} & \textbf{R} & \multicolumn{1}{c}{\textbf{P}} & \multicolumn{1}{c}{\textbf{F1}} \\ \hline
\textit{\textsc{lookup}} & \multicolumn{1}{c}{0.386} & \multicolumn{1}{c}{0.555} & \multicolumn{1}{c|}{0.455} & \multicolumn{1}{c}{0.591} & 0.849 & 0.697 \\
\textit{\textsc{noun} (gold tags)} & \multicolumn{1}{c}{0.123} & \multicolumn{1}{c}{0.111} & \multicolumn{1}{c|}{0.117} & \multicolumn{1}{c}{0.855} & 0.773 & 0.812 \\
\textit{\textsc{noun} (pred tags)} & 0.121 & 0.107 & 0.113 & 0.853 & 0.756 & 0.802 \\
\textit{\textsc{parse} (gold parse)} & 0.879 & 0.862 & \textbf{0.870} & 0.948 & 0.929 & \textbf{0.938} \\
\textit{\textsc{parse} (predicted)} & 0.831 & 0.815 & \textbf{0.823} & 0.941 & 0.922 & \textbf{0.931} \\
\textit{\textsc{sequence} (10)} & 0.463 & 0.651 & 0.541 & 0.611 & 0.859 & 0.714 \\
\textit{\textsc{sequence} (binary)} & 0.653 & 0.732 & 0.690 & 0.793 & 0.725 & 0.757
\end{tabular}
\caption{\label{tab:mention} Results for automatic entity mention span detection, exact span match on the left and fuzzy match containing entity heads on the right.}
\end{table*}

Table \ref{tab:mention} gives scores for baselines (\textsc{noun} and \textsc{lookup}) and the competitive approaches (\textsc{parse} and \textsc{sequence}). For \textsc{noun} and \textsc{parse} we provide separate scores for gold versus predicted POS and trees, using Marmot \cite{MuellerSchmidSchuetze2013} as the tagger, and MaltParser \cite{Nivre2009} for parsing. For \textsc{sequence}, we tested two scenarios: 10-way classification (typed entities) and binary classification (entity/non-entity), which should be easier to learn given the limited data. In all cases, metrics evaluate correct/incorrect boundary detection, ignoring entity types. Fuzzy span scores are more lenient, matching spans that include the entity's lexical head word, even if exact boundaries are incorrect. The results confirm the suspicion that training data and/or word embedding representations are insufficient for high quality results using the neural system. The binary RNN does better by 15\%, suggesting training data sparseness may be the main issue. \textsc{parse} methods are promising, indicating that investment in a higher quality parser may help. For fuzzy span scores, the small degradation of the parse-based strategy and \textsc{noun} using automatic NLP is due to the fact that tagging nouns in Coptic is comparatively easy, especially if we ignore the common vs. proper nouns distinction (both of which usually indicate mentions equally).

\paragraph{Entity classification} As spans for entity classification evaluation, we use automatic parser output for all strategies, except for \textsc{sequence}, which uses the RNN's predicted spans. Scores are assigned for both exact match (span and entity type) and fuzzy match (minimal span containing the head receiving the correct entity type). As a baseline, we select the majority class \textsc{abstract} for all spans identified by the parser. As a third approach, we apply Knowledge Base (\textsc{kb}) lookup. To create our KB, we annotated the most frequent 2,700 nouns from Coptic Scriptorium with possible entity types regardless of context; however to keep the experimental setup fair, we use a version of the KB with only those lexical items which are attested in the training set (about 1,300 entries). We note this approach cannot handle novel words which are not included in the KB (for these we guess the majority \textsc{abstract}), and has no way of disambiguating ambiguous entries (for which we guess the attested majority category from training). 

Finally, we test two feature-based approaches, in which a conditional random fields (CRF) model is adopted, using scikit-learn's CRF Suite.\footnote{A reviewer has asked why a CRF classifier is useful when we cannot use BIO encoding for nested span detection. In fact, in our experiments CRFs outperformed other word-wise classifiers, such as Random Forest and Gradient Boosting, since they can constrain transitions between labels which are helpful even for head word classification. Adjacent words may have plausible but incompatible tags (e.g.~for the saint `Apa Shenoute', both words can be \textsc{person}, but only one should be labeled as the head), and certain transitions, such as inanimate object followed by an animate possessor, can be captured by CRFs.} The model takes selected features as input, and outputs predicted entity type for each entity's head token position only, taking into account the most probable path of labels through each sentence. Three categories of features are extracted from the input data:

\begin{itemize}[itemsep=0pt]
    \item \textbf{Grammatical features:} 1. first/last 2-3 characters of each token, giving access to some morphological affixes (e.g.~initial \textit{mnt-} forms abstracts, like English `-ness'); 2. POS tags and dependency functions; 3. syntactic parent: for example subjects of verbs like `say' are likely to be a \textsc{person}.
    \item \textbf{Numerical features:} 1. descendent span length, i.e.~how many words are dependent on the current token, directly or indirectly. 2. percentile position in sentence: humans are often mentioned earlier in sentences, whereas inanimate modifiers tend to occur late; 3. sentence length.
    \item \textbf{Context features:} previous and next tokens and their POS tags and dependency functions.
\end{itemize}

Beyond testing the CRF classifier as a standalone solution, we also combine it with the KB resources. In this setup, input items found in the KB are classified according to their entries, and the CRF classifier is only consulted in three scenarios: 

\begin{enumerate}[itemsep=0pt]
    \item when there are out of vocabulary (OOV) tokens for the KB (i.e.~unknown words)
    \item when an item has multiple KB entries, we choose the one with the higher CRF classifier score
    \item when the CRF classifier is highly confident that an item is non-referential (i.e.~predicting the `O' class with probability \textgreater 95\%), the entity candidate is discarded
\end{enumerate}

\begin{table*}[h!bt]
\centering
\begin{tabular}{l|cll|cll}
 & \multicolumn{3}{c|}{\textbf{span match}} & \multicolumn{3}{c}{\textbf{head match}} \\
\textbf{method} & \textbf{R} & \multicolumn{1}{c}{\textbf{P}} & \multicolumn{1}{c|}{\textbf{F1}} & \textbf{R} & \multicolumn{1}{c}{\textbf{P}} & \multicolumn{1}{c}{\textbf{F1}} \\ \hline
\textit{\textsc{majority}} & 0.213 & \multicolumn{1}{c}{0.209} & \multicolumn{1}{c|}{0.211} & 0.235 & 0.230 & 0.232 \\
\textit{\textsc{sequence}} & 0.476 & \multicolumn{1}{c}{0.614} & \multicolumn{1}{c|}{0.536} & 0.527 & 0.757 & 0.621 \\
\textit{\textsc{kb}} & \multicolumn{1}{l}{0.681} & 0.660 & 0.670 & \multicolumn{1}{l}{0.728} & 0.705 & 0.717 \\
\textit{\textsc{crf}} & \multicolumn{1}{l}{0.805} & 0.778 & 0.791 & \multicolumn{1}{l}{0.861} & 0.831 & 0.846 \\
\textit{\textsc{crf+kb}} & \multicolumn{1}{l}{\textbf{0.827}} & \textbf{0.810} & \textbf{0.818} & \multicolumn{1}{l}{\textbf{0.889}} & \textbf{0.869} & \textbf{0.879}
\end{tabular}
\caption{\label{tab:classification} Scores for entity type identification. All methods except RNN use spans predicted by the best method for mention detection.}
\end{table*}

Table \ref{tab:classification} provides the results of all models. The RNN model (\textsc{sequence}) is not competitive, probably due to the limited size of training data and word embeddings. KB gains approx. 14\%, showing it covers many more cases. For the CRF model, F1 scores rise to 0.79 and 0.81, indicating the feature-based model is promising. The hybrid approach (\textsc{crf+kb}) performs best, due to the ability of the CRF classifier to disambiguate uncertain cases and the KB's power to capture rare and unambiguous items (e.g.~less frequent categories such as \textsc{plant} or \textsc{event}, which the CRF dismisses as unlikely).

\paragraph{Wikification} For entity linking, our data is too small to use neural approaches with word embedding inputs. Due to the limited training data, many relevant identities will not appear in our data, meaning a large part of the target values for linking are unknown for our system. At present we therefore use a rudimentary semi-automatic strategy, offering possible links to human annotators, who can accept or reject suggestions, and enter new links for entities that appear for the first time. Our lookup strategy uses a heuristically ordered cascade applied to all minimal spans containing a proper noun:

\begin{itemize}[itemsep=0pt]
    \item If the exact entity text is known in other documents in the same corpus, prefer the most frequent link associated with it (e.g. `John who gives baptism' in the Gospel of Mark is `John the Baptist')\footnote{This is relevant e.g.~because the Treebank includes only chapters 1--9 of Mark, but we want to annotate entire works.}
    \item Otherwise, if the entity text has appeared elsewhere in the corpus, prefer its most frequent link (exact match for `John who gives baptism' is better, even if the corpus has other more frequent Johns)
    \item Else, if the entity's head noun is known in this corpus, prefer its most frequent link (`John' in the Gospel of Mark is most often `John the Baptist')
    \item Else, if the entity's head is known anywhere, prefer its most frequent link (`John' might overall most frequently refer to `John the Apostle' in all sub-corpora)
\end{itemize}

This cascaded heuristic can only work for proper nouns that appear somewhere in the training data, and is susceptible to a majority bias  (i.e.~it always guesses that a `John' in a new corpus is the most frequent John in our data).

To evaluate our strategy, we compare it with two baselines: exact match majority choice (most frequent link matching the entire entity string) and head match (most frequent entity associated with a head noun). We use both the train and dev partitions to build our lookup table, while the test set remains the same, containing some 100 identifiable entity mentions belonging to 53 distinct types.

\begin{table*}[h!bt]
\centering
\begin{tabular}{l|ccc}
\multicolumn{1}{c|}{\textbf{method}} & \textbf{acc} & \textbf{cov} & \textbf{no\_err} \\ \hline
\textit{exact} & 0.227 & 0.273 & 0.953 \\
\textit{head} & 0.433 & 0.500 & 0.933 \\
\textit{cascade} & \textbf{0.460} & \textbf{0.500} & \textbf{0.960}
\end{tabular}
\caption{\label{tab:linking} Wikification scores -- accuracy (\% correct links), coverage (\% entities for which a response is retrieved), and \% entities with no false links (correct link matched, or entity not covered).}
        \par\vspace{-10pt}\par
\end{table*}

Table \ref{tab:linking} shows that the cascade improves on the baselines, and that, while coverage is limited (only 50\%), it rarely misclassifies an example. Errors arise due to single word names shared between multiple entities, such as John (the Apostle or the Baptist) or Paul (the Apostle, or Paul of Thebes), and context differences, such as  `Israel', which is linked to `Kingdom of Israel (united monarchy)' when referring to King David's kingdom, but to `Israelites' when referring to Israel as a people. We currently feed such predictions to human annotators for disambiguation.

\section{Applications}\label{sec:applications}

\paragraph{Distant reading} Entity information can help address a variety of research questions, as well as making data more accessible and easier to discover in the larger DH ecosystem \cite{Schroeder2020}. As a first way of looking at entities in Coptic text, we can consider how to visualize the global picture of entity frequencies in our data as a kind of distant reading \cite{Moretti2013}. Two interactive visualizations we can use for this purpose are entity term networks and recursive TreeMaps \cite{Shneiderman1992}.

\begin{figure}[h]
        \centering
 \begin{subfigure}{.475\textwidth}
            \includegraphics[width=\textwidth,height=110pt,frame]{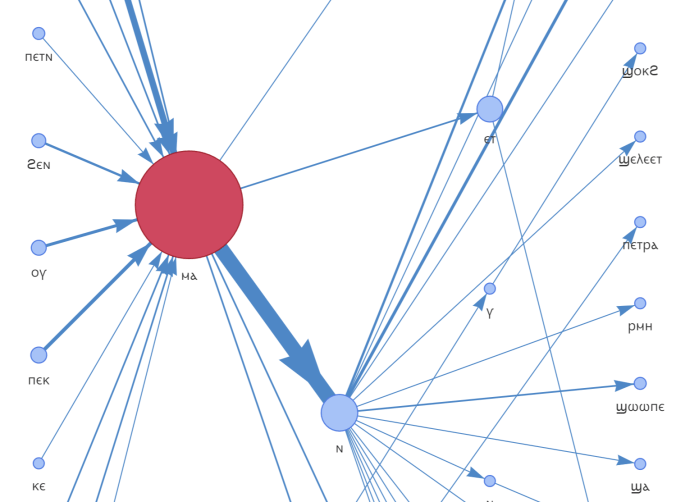}
            \caption[]{Entity Network for \begin{coptic}ma\end{coptic} \textit{ma} `place.'}\label{fig:network}
        \end{subfigure}\hfill
               \begin{subfigure}{.475\textwidth}
            \includegraphics[width=\textwidth,frame]{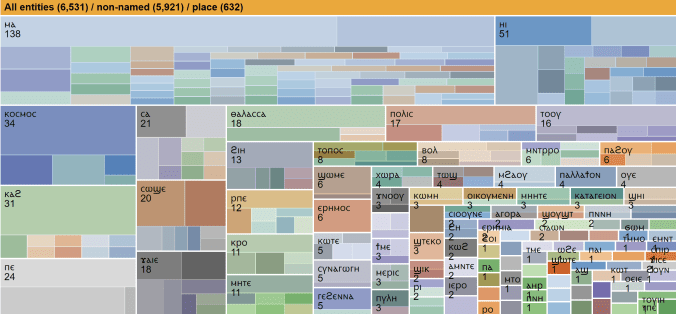}
            \caption[]{TreeMap of non-named place entities}\label{fig:treemap}
        \end{subfigure}
        \caption{Distant reading visualizations.}
        \par\vspace{-10pt}\par
    \end{figure}

Entity term networks visualize a head word's relationships with other words in its entity spans. Figure \ref{fig:network} captures part of the network for \begin{coptic}ma\end{coptic} \textit{ma} `place'. Larger nodes represent more frequently recurring terms, and broad arrows correspond to frequent transitions. The network for \begin{coptic}ma\end{coptic} gives us a clearer idea of its potential semantic relationships: often preceded by \begin{coptic}pek\end{coptic}/\begin{coptic}petn\end{coptic} \textit{pek/petn} `your', almost always followed by \begin{coptic}n\end{coptic} \textit{n} `of', continuing to nouns indicating purpose (`place of dwelling', or `lavatory' with \begin{coptic}rmh\end{coptic} \textit{rmē} `urination'), events (\begin{coptic}yeleet\end{coptic} \textit{šeleet} `wedding'), directions (\begin{coptic}ya\end{coptic} \textit{ša} `East') and more. Similarly, the TreeMap concisely depicts the most mentioned entities and headwords in the corpora. Following our annotation guidelines, the TreeMap initially separates named and non-named entities. These are divided into the ten entity types, and then by unique entity head words. Figure \ref{fig:treemap} shows the view of non-named place entities. The TreeMap shows that places headed by \begin{coptic}ma\end{coptic} (top left, 138 cases) are most frequent, followed by \begin{coptic}hi\end{coptic} \textit{ēi} `house' (51) and \begin{coptic}kocmoc\end{coptic} \textit{kosmos} `cosmos, world' (34). In lower ranks, the data evinces the importance of the desert in our corpus. Many texts center on monks who have left civilization, and their setting is often the \begin{coptic}djaie\end{coptic} \textit{jaie} `desert' (18). Consequently, we see our texts mention \begin{coptic}djaie\end{coptic} more than \begin{coptic}polic\end{coptic} \textit{polis} `city' (17), and even more so if we include Greek synonyms such as \begin{coptic}erhmoc\end{coptic} \textit{erēmos} `desert' (6). These visualizations allow easy interactive exploration and show patterns that may be missed when reading individual texts. Without entity annotation, such phrases cannot be trivially extracted and categorized for comparison.

\paragraph{Entity type proportions}

Another comparable quantity across texts is the proportion of named vs. non-named entities, or proportions of entity types. For the latter, we observe that sermons in the treebank (e.g.~Pseudo-Athanasius, Letters of Besa) have much more abstract entities than narratives, since they concentrate on instruction, using abstractions to communicate their message and mentioning people less often. Narratives mention people more frequently, with a person/abstract ratio of 2:1 or higher, compared to $\sim$1:1 in sermons. Drilling down im more detail, we find exceptions such as The Life of Onnophrius and The Dormition of John: these have person/abstract ratios close to 1:1, mainly due to homiletic speeches delivered by the main characters, echoing Coptic sermons as they instruct disciples to avoid sin. This quantitative finding foregrounds an interesting commonality between seemingly unrelated texts and differences between texts from one genre.

We imagine many more findings will emerge from Coptic entity annotations, which complement detailed philological, literary, and historical inquiries. Data aggregation and visualization of different subsets of texts enable analyses based on the quantity, proportion and dispersion of entity types which we are only beginning to explore. They abstract away from individual ways of phrasing references to people and places, while linking mentions of named entities across datasets, projects, and DH tools.

\section{Summary and conclusion}\label{sec:conclusion}

In this paper we presented a new annotated data set for Coptic entity classification and linking, using ten entity types, including named and non-named, potentially nested entities, and attaching named entities to corresponding Wikipedia articles, i.e.~Wikification. Our annotated data represents a wide range of genres, including translated and autochthonous texts, and is freely available in a number of popular formats, including the CoNLL and TEI XML formats, under an open license, as are our tools.\footnote{Available at: \url{https://github.com/CopticScriptorium/corpora}, including automatically annotated silver data. Gold entity annotations have also been merged into the release of the UD Coptic Treebank, at \url{https://github.com/UniversalDependencies/UD_Coptic-Scriptorium}.}

From a technical perspective, our results demonstrate the difficulty in applying state-of-the-art neural frameworks to nested NER in languages with modest data sizes. At the same time, the lack of availability of many millions of words for training word embeddings limits the utility of RNN architectures relying on highly informative, context sensitive information. Instead, we are able to show that a syntax-based approach using dependency trees to identify nested noun phrases is viable and substantially more accurate for data in the several 10K-100K tokens range. Although our approach relies on the existence of such syntactically annotated data to train a parser, the size of the data set is likely to be a more realistic target for projects in similar settings to ours than the size of datasets underlying standard approaches to (N)NER in modern languages. For comparison, the Universal Dependencies treebanks for Ancient Greek and Latin include over 400,000 and 800,000 tokens respectively, meaning that the treebank used here is very modestly sized; and even with larger treebanks, those languages too lack highly expressive word embeddings based on hundreds of millions, or even billions of words, which are commonly available for modern European languages. Our methods therefore show promise for mention detection in annotating other resources for classical languages, such as freely available Ancient Greek and Latin texts available in the Perseus Digital Library \cite{SmithEtAl2000}.

For entity classification, our approach shows the robustness of feature based classifiers (Section 3), while also revealing the added value of a knowledge base (KB) architecture. A KB derived strictly from the training data is already helpful for the hybrid approach taken here (KB+CRF in Section 3), while an even broader coverage KB can be constructed with relatively low effort, which makes it possible to capture some of the rarer, but often lexically unambiguous entity types, such as names of animals or plants. For entity linking, we show modest results in terms of recall, but quite good precision which can facilitate larger manual annotation efforts by offering reliable suggestions for human review.

In Section 4 we outlined some of the applications that wide-coverage entity annotations can bring for humanists, including style and genre studies, highlighting differences between documents that are otherwise similar from the text type perspective, and bird's eye-view visualizations which allow us to examine how texts talk about entities, and how often. The data used for the example studies in this paper comes from the high quality, but small, manually annotated corpus prepared for this work. We plan to publish a much larger automatically annotated corpus of Coptic texts using the tools described here, which we expect to deliver much more comprehensive capabilities in exploring the contents of Coptic texts, many of which still have no translations that might allow exploring the content in English. Such larger scale data could also be used to link to other projects featuring entity identifiers (in Coptic or otherwise), and to lexicographic projects, such as the Coptic Dictionary Online \cite{FederKupreyevManningEtAl2018}, or the Database and Dictionary of Greek Loanwords in Coptic (DDGLC) \cite{AlmondHagenJohnEtAl2013}.

\section{Acknowledgments}

We would like to thank Coptic Scriptorium contributors Mitchell Abrams, Elizabeth Davidson, Rebecca Krawiec, Christine Luckritz Marquis, Elizabeth Platte, Dana Robinson and Caroline T. Schroeder for their work on the annotated data, as well as the anonymous reviewers for their feedback. This work was supported by a Stage III Digital Humanities Advancement Grant from the National Endowment for the Humanities (grant HAA-261271-18). 

\bibliographystyle{coling}
\bibliography{coling2020}

\vspace{1cm}

\pagebreak

\end{document}